\documentclass[10pt,twocolumn,letterpaper]{article}

\usepackage[pagenumbers]{cvpr} 

\definecolor{cvprblue}{rgb}{0.21,0.49,0.74}
\usepackage[pagebackref,breaklinks,colorlinks,allcolors=cvprblue]{hyperref}

\newcommand{\OURS}{TeamHOI}

\newcommand{\boldparagraph}[1]{\vspace{0.05cm}\noindent{\bf #1:} }

\usepackage{epsfig}
\usepackage{graphicx}
\usepackage{amsmath}
\usepackage{amssymb}
\usepackage{caption}
\usepackage{multirow}
\usepackage{url}
\usepackage{xcolor}
\usepackage{wrapfig}
\usepackage{algorithm2e}
\usepackage{comment}
\usepackage{hyperref}

\usepackage{float}

\usepackage{booktabs}
\usepackage{lipsum} 

\usepackage{array}       
\usepackage{booktabs}    
\usepackage{siunitx}

\def\paperID{*****} 
\def\confName{CVPR}
\def\confYear{2026}

\title{\OURS: Learning a Unified Policy for Cooperative Human-Object Interactions with Any Team Size}

\author{%
Stefan Lionar$^{1,2,3}$ \qquad Gim Hee Lee$^{3}$ \vspace{1pt}\\
$^{1}$Garena \qquad\quad $^{2}$Sea AI Lab  \qquad\quad $^{3}$National University of Singapore \vspace{8pt} 
}

\begin{document}

\twocolumn[{%
        \maketitle
	\begin{center}
    \vspace{-11mm}
        {\small \url{https://splionar.github.io/TeamHOI}}
        \vspace{2mm}
		
        \includegraphics[width=0.98\linewidth]{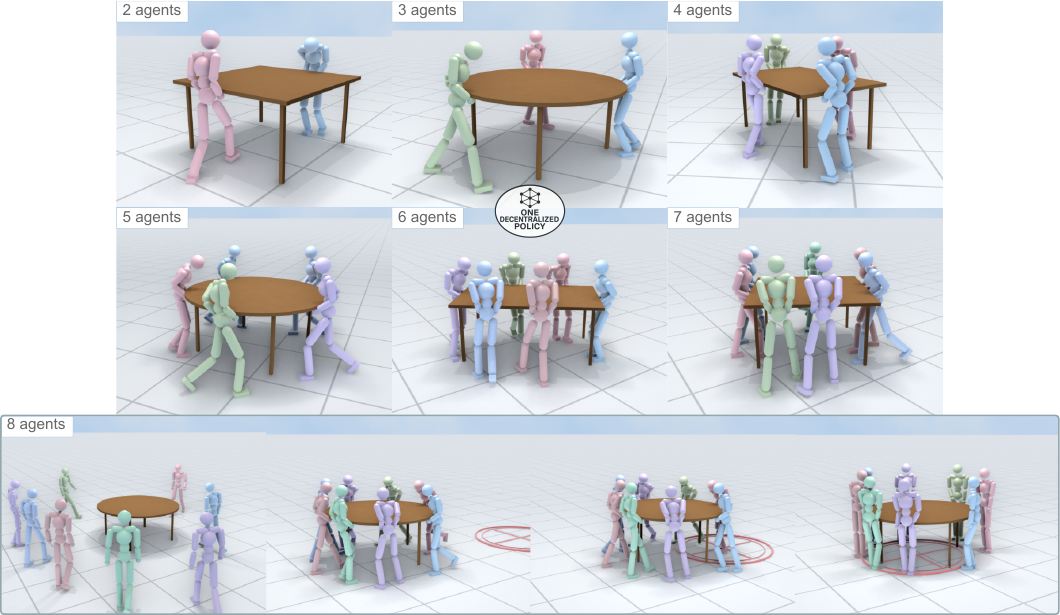}
    \captionsetup{hypcap=false}\captionof{figure}{We present \textbf{\OURS{}}, a framework for learning a unified decentralized policy for cooperative human-object interactions (HOI) across varying team sizes and object configurations. Our framework enables effective cooperation where each humanoid acts independently from local observations while coordinating with others through a single shared policy. Video demonstrations are provided on our \href{https://splionar.github.io/TeamHOI}{webpage}.}
		\label{fig:teaser}
	\end{center}    
}]

\maketitle

\begin{abstract}

Physics-based humanoid control has achieved remarkable progress in enabling realistic and high-performing single-agent behaviors, yet extending these capabilities to cooperative human-object interaction (HOI) remains challenging. We present \OURS{}, a framework that enables a single decentralized policy to handle cooperative HOIs across any number of cooperating agents. Each agent operates using local observations while attending to other teammates through a Transformer-based policy network with teammate tokens, allowing scalable coordination across variable team sizes. To enforce motion realism while addressing the scarcity of cooperative HOI data, we further introduce a masked Adversarial Motion Prior (AMP) strategy that uses single-human reference motions while masking object-interacting body parts during training. The masked regions are then guided through task rewards to produce diverse and physically plausible cooperative behaviors. We evaluate \OURS{} on a challenging cooperative carrying task involving two to eight humanoid agents and varied object geometries. Finally, to promote stable carrying, we design a team-size- and shape-agnostic formation reward. \OURS{} achieves high success rates and demonstrates coherent cooperation across diverse configurations with a single policy.

\end{abstract}

\section{Introduction}
\label{sec:intro}

Physics-based humanoid control and human-object interaction (HOI) have rapidly advanced, enabling virtual humans and robots to walk, grasp, and manipulate objects with realistic motion~\cite{gu2025humanoid,sui2026survey,wang2025physhsi}. Yet many everyday tasks, such as lifting large and heavy items, require multiple agents to coordinate their physical actions. Building humanoids that can cooperate in such settings is a key step toward more capable and intelligent systems. Beyond robotics, this ability also opens up exciting directions for next-generation creative AI applications, such as multi-character animation and interactive game worlds, where virtual humanoids must coordinate naturally with each other.

However, existing physics-based humanoid motion frameworks still face major limitations in both \emph{scalability} and \emph{data diversity} when applied to cooperative HOI. Most approaches rely on fixed-size input MLP policies to generate control actions, and employing such architectures for multi-agent interactions restricts the policy to a fixed team size~\cite{luo2024smplolympics}. Another method omits explicit agent-to-agent communication altogether, and instead relying solely on shared object dynamics as an indirect communication channel~\cite{gao2024coohoi}. Such designs fail to capture the adaptive nature of real human cooperation, where individuals continuously perceive their teammates’ presence and adjust their coordination according to the team’s composition and size.

Another key limitation lies in the data source. Many physics-based HOI frameworks leverage the \emph{Adversarial Motion Prior (AMP)} to ensure that learned motions remain natural by regularizing them toward reference motion data. However, reference motion for coordinated multi-human activities is mostly unavailable, necessitating cooperative HOI frameworks to rely on single-human demonstrations. This restriction limits the achievable cooperative behavior. The coordination patterns can only be tied to the motion of one demonstrator, reducing flexibility when handling cooperative HOI with larger groups of agents.

To address these limitations, we propose \emph{\OURS{}}, a framework that enables a single decentralized policy to generalize across any number of cooperative agents. Each agent operates independently using local observations, while sharing the same policy network parameter. To model cooperation with flexible team size, we employ Transformer-based policy network, effectively removing the fixed-size input restriction from MLP policy, and incorporate the states of other agents as \emph{teammate tokens}. The policy is trained in environments instantiated with different team size configurations, exposing it to diverse coordination patterns and allowing it to adapt seamlessly to varying team sizes with their corresponding coordination demands without retraining or fine-tuning.

\OURS{} also addresses the data diversity limitation associated with motion priors. To expand the diversity of feasible cooperative behaviors, we use reference motions from single human actor while masking out the body parts interacting with objects during AMP supervision. We then enforce the masked regions to achieve desirable interactions through task rewards. For example, a sideways walking reference motion can be repurposed for sideways lifting by adapting the hand-object interaction reward. This masking strategy effectively broadens the range of feasible HOI skills, enabling diverse cooperative behaviors to emerge from single-human reference motions.

As a concrete testbed to evaluate our proposed framework, we demonstrate \OURS{} through a challenging cooperative carrying task, as illustrated in Figure~\ref{fig:teaser}. In this task, a team of humanoid agents must approach and transport an object, specifically a table that can come in either square, rectangular, or round shape. The agents have to coordinate to establish formations that promote stable lifting, and collectively transport the table to a desired target location. To accomplish this task, we design a \emph{formation reward} that is agnostic to both the table shape and the number of cooperating agents, guiding the agents to distribute themselves into stable positions for carrying. Through extensive experiments, we show that our proposed framework enables a single policy to perform seamlessly across configurations with two to eight cooperating agents, achieving both high success rates and coherent cooperative behaviors.

In summary, our contributions are as follows: 
\begin{itemize}
    \item We introduce \emph{\OURS{}}, a framework that enables a single decentralized policy to perform cooperative human-object interaction with any number of agents.
    \item We employ a Transformer-based policy network that learns from diverse team-size configurations and adapts the required coordination through teammate tokens.
    \item We propose a masked AMP strategy that overcomes the data diversity limitation in previous motion-prior methods, expanding the range of feasible and diverse cooperative HOI behaviors.
    \item We demonstrate the emergent cooperative behaviors of \OURS{} in challenging table-carrying tasks involving varied object shapes and team sizes.
    \item We design a formation reward that promotes stable carrying, agnostic to both the table shape and the number of cooperating agents.
\end{itemize}

\section{Related Work}
\label{sec: related_work}

\subsection{Physics-based Human-Scene Interaction}
Physics-based human-scene interaction (HSI) focuses on enabling humanoid agents to interact with objects and environments under realistic physical conditions, including contact, friction, and dynamic constraints. These interactions are typically realized through physics-based control in modern simulators such as Isaac Gym~\cite{makoviychuk2021isaac} and MuJoCo~\cite{todorov2012mujoco}. A common training paradigm is reference tracking with deep reinforcement learning (RL)~\cite{liu2015improving,peng2018deepmimic,chao2021learning,xie2023hierarchical,merel2020catch,zhang2023simulation}, often enhanced by Adversarial Motion Priors (AMP)~\cite{peng2021amp}, whose style reward aligns synthesized motions to the reference motions. AMP has proven effective in improving motion realism across diverse downstream tasks~\cite{juravsky2022padl,peng2022ase,tessler2023calm} and HSI specifically~\cite{hassan2023synthesizing,pan2024synthesizing,xiao2023unified}. This line of works has led to a broad range of contact-rich skills, such as in sports environments~\cite{zhang2023vid2player3d,wang2024strategy,luo2024smplolympics,liu2018learning,wang2025skillmimic,yu2025skillmimic, liu2017learning}, and everyday object interactions~\cite{hassan2023synthesizing,xiao2023unified,pan2024synthesizing,li2024physics,wang2024sims,gao2024coohoi,li2025learning,liu2024mimicking,braun2024physically,luo2024omnigrasp,wu2025human,xu2025intermimic,xu2026interprior,tessler2025maskedmanipulator}.

Beyond mastering individual skills, recent efforts seek broader and more reusable capabilities. Motion-manifold and skill-library approaches learn versatile priors that can be composed across tasks~\cite{peng2022ase,dou2023c,bae2023pmp,serifi2024vmp,huang2025modskill}, while unified controllers aim to capture diverse behaviors within a single policy~\cite{luo2023universal,tessler2024maskedmimic,wu2025uniphys,luo2023perpetual,won2019learning,won2020scalable,he2025hover}.
More recently, TokenHSI~\cite{pan2025tokenhsi} introduces task tokenization with a Transformer-based policy~\cite{vaswani2017attention}, enabling multi-skill unification for versatile HSI and flexible adaptation to new tasks.

\subsection{Multi-Humanoid Interaction and Cooperation}
Research on multi-humanoid interactions remains relatively limited compared to single-agent motion synthesis. Several multi-humanoid datasets have been introduced, ranging from everyday human-human activities~\cite{kundu2020cross,shen2019interaction,fieraru2020three,xu2024inter, liu2025core4d, zhang2024hoi, kogashi2026mmhoi} to choreographic multi-person motions~\cite{le2023music,siyao2024duolando}. Building on these data sources, numerous kinematic-based multi-character animation methods have been proposed~\cite{won2014generating,liang2024intergen,tanaka2023role,shafir2023human,ghosh2024remos,tan2025think,zhang2025reactffusion,zhao2025freedance,fan2024freemotion,ghosh2025duetgen,javed2024intermask}. Although visually compelling, these approaches rely heavily on high-quality interaction data and cannot guarantee physical plausibility.

To move beyond purely kinematic synthesis, recent works explore physics-based multi-character interactions using kinematic generators~\cite{huang2025diffuse, tevet2024closd, karunratanakul2023guided, zhang2024tedi} and PHC~\cite{luo2023perpetual}, which converts multi-human kinematic motion into state-action pairs for policy learning in physics simulation. This enables interactive physical behaviors among multiple agents~\cite{liu2024physreaction,ji2025towards,li2025interagent}, albeit without object. Other physics-based efforts focus on crowd navigation~\cite{haworth2020deep,rempe2023trace} that model collision avoidance and group motion. 

Another line of works has incorporated object interactions into the multi-humanoid settings. An imitation-based approach~\cite{zhang2023simulation} demonstrates multi-character interactions involving a shared object, but remains constrained by the scarcity of high-quality multi-human motion capture. Meanwhile, SMPLOlympics~\cite{luo2024smplolympics} showcases multi-humanoid sports behaviors in physics simulation, yet operates with fixed small team sizes. Most relevant to our work is CooHOI~\cite{gao2024coohoi}, which models cooperative human-object interaction by relying on implicit communication through shared object dynamics. However, it does not incorporate the states of other agents, a limitation that is unrealistic for human cooperation where agents continuously perceive and respond to one another. Moreover, its dependence on full-body single-actor reference motions limits the diversity of achievable cooperative behaviors. For instance, the agents are shown to only perform forward and backward lifts and cannot adapt to more varied cooperative HOI strategies.

\section{Methodology}

Our goal is to develop a unified decentralized policy for cooperative HOI that generalizes across varying team sizes and their corresponding coordination demands. Each agent acts independently based on local observations while incorporating the states of other agents for effective cooperation through a single policy.

\subsection{Preliminary}

\begin{figure*}[ht!]
  \centering
   \includegraphics[width=0.82\linewidth]{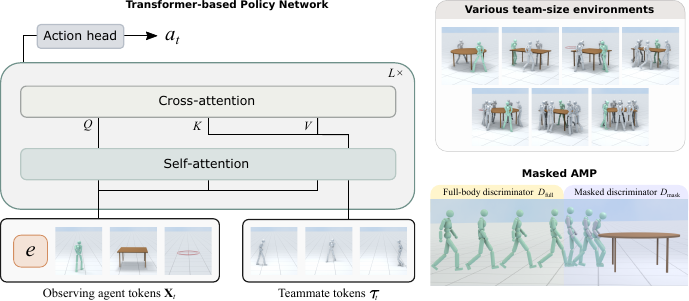}
   \caption{\textbf{Overview of \OURS{} framework.}
A transformer-based policy network enables coordination between the observing agent (green humanoid) and its teammates (grey humanoids) through alternating self- and cross-attention layers. By training across diverse team-size environments, the framework learns a unified policy that works across different team configurations. To maintain motion realism and enhance skill diversity, a masked AMP strategy blends full-body and masked discriminators based on object interaction.}
   \label{fig:pipeline}
   \vspace{-0.3cm}
\end{figure*}

We build on the AMP framework~\cite{peng2021amp}, which augments reinforcement learning with motion prior that enforces motion realism.
In AMP, a policy $\pi_\theta$ is trained together with a discriminator $D_\phi(s,s')$ that distinguishes short state transitions $(s,s')$ from reference motion data versus those generated by the policy.
The discriminator provides a style-based feedback signal that encourages the policy to produce realistic motion transitions.

\boldparagraph{RL Setup}
Each agent observes a proprioceptive state $s_t$ (joint angles, velocities, and root pose), an optional goal state $g_t$ (e.g., object target position), and outputs an action $a_t$ (target joint positions or torques).
The environment evolves according to the simulator dynamics to produce the next states and a task reward $r_t^{\text{task}}$.
The policy $\pi_\theta(a_t|s_t, s_g)$ is optimized using Proximal Policy Optimization (PPO)~\cite{schulman2017proximal}, maximizing the expected discounted return: $J(\theta) = \mathbb{E}_{\pi_\theta}\!\left[\sum_t \gamma^t r_t \right]$.

\boldparagraph{Style Reward}
To incorporate motion realism, AMP introduces an additional style reward from the discriminator:
\begin{equation}
    r_t^{\text{style}} = - \log (1 - D_\phi(s, s')),
\end{equation}
where $(s, s')$ denotes the current and next states, capturing short-term motion dynamics.
The total reward combines both terms as:
\begin{equation}
    r_t = r_t^{\text{task}} + \lambda_{\text{AMP}}\, r_t^{\text{style}},
\end{equation}
where $\lambda_{\text{AMP}}$ balances task performance and motion realism.
The policy is optimized via the PPO objective using $r_t$, while the discriminator is trained to classify transitions from the policy and those from the reference dataset:
\begin{align}
    \mathcal{L}_D
    &= -\,\mathbb{E}_{(s,s')^{\text{ref}}}[\log D_\phi(s,s')] \nonumber \\
    &\quad -\,\mathbb{E}_{(s,s')^{\pi}}[\log (1 - D_\phi(s,s'))].
\end{align}

\subsection{\OURS{} Framework}

Our \OURS{} framework (Figure~\ref{fig:pipeline}) reformulates AMP within a flexible multi-agent reinforcement learning setup that scales to an arbitrary number of humanoid agents.

\boldparagraph{Policy Network}  
To enable scalable coordination, we employ a Transformer-based architecture as our policy network, inspired by TokenHSI~\cite{pan2025tokenhsi}. Each humanoid agent obtains observation $o_t \triangleq (s_t, g_t)$, which consists of its proprioceptive state $s_t$ and goal states $g_t$. Each observation component is first processed by a dedicated tokenizer to produce the token sequence of the main observing agent,
$\mathbf{X}_t = [\, e,\, \mathbf{T}_t^{s},\, \mathbf{T}_t^{g} \,]$,
where $\mathbf{T}_t^{s}$ and $\mathbf{T}_t^{g}$ denote the proprioceptive and goal tokens, and $e$ is a learnable embedding preceding the action head.

To enable coordination with variable team sizes, the observing agent’s policy attends to a set of \emph{teammate tokens} $\{\mathcal{T}_t^{i}\}_{i=1}^{N-1}$, each encoding the cues of another agent (e.g., position, heading direction) expressed in the observing agent’s local frame. The transformer backbone consists of $L$ stacks of alternating self-attention and cross-attention layers. Self-attention operates over the observing agent tokens $\mathbf{X}_t$, while cross-attention enables these tokens to attend to the teammate tokens efficiently even when the team size is large. The updated embedding $e$ is passed through an action head to predict control output $a_t$. The overall policy is defined as $a_t = \pi_\theta(a_t \mid \mathbf{X}_t, \{\mathcal{T}_t^{i}\}_{i=1}^{N-1})$, where the control output corresponds to the target joint rotation for each actuated degree of freedom for PD controller.

\boldparagraph{Training a Unified Policy}  
Within RL framework, each environment represents an independent simulation instance where agents interact with the world, observe states, take actions, and receive rewards. To learn a single policy that generalizes across different collaboration scenarios, we instantiate multiple environments in parallel, each configured with a different number of cooperating agents and their distinctive cooperative rewards. Through this setup, the policy network is trained on diverse multi-agent configurations, gaining exposure to varying interaction dynamics across team sizes. To ensure stable training across mixed team configurations, we normalize PPO advantages separately for each team size. Further details are provided in the supplementary material.

\boldparagraph{Masked AMP}  
A major challenge in extending AMP to cooperative HOI is the lack of multi-human reference motion data. Although single-human reference motions can be used, directly regularizing the policy toward them limits the diversity of cooperative behaviors that can emerge, as cooperative tasks often require a wider range of locomotion skills than those present in a single demonstrator.

To address this limitation, we introduce a \emph{Masked AMP} strategy that maintains style realism while allowing diverse HOI behaviors. Specifically, two discriminator networks are trained: one full-body AMP network $D_\text{full}$ that evaluates complete reference motion, and one masked AMP network $D_\text{mask}$ that excludes body parts directly interacting with the object (e.g., hands and forearms). During object interaction, the style reward \(r_t^{\text{mask}}\) is derived from \(D_{\text{mask}}\), whereas \(r_t^{\text{full}}\) from \(D_{\text{full}}\) is applied when the humanoid is not interacting with the object. The overall blended style reward is: 

\vspace{-0.2cm}
\begin{equation}
r_t^{\text{style}} = \sigma(\alpha_t)\, r_t^{\text{mask}} + (1 - \sigma(\alpha_t))\, r_t^{\text{full}},
\end{equation}
where $\sigma$ is a sigmoid function operating on continuous interaction indicator $\alpha_t$ (e.g., agent-object distance). 

Our formulation shares conceptual similarities with part-wise motion priors (PMP) framework~\cite{bae2023pmp}, which assembles motion skills from different body segments to enrich physical interaction. However, unlike PMP which learns part-wise priors directly, our method enforces diversity in the masked regions through task rewards to enable adaptable object interactions from limited single-body references.

\subsection{Cooperative Carrying Task}\label{sec:cooptask}

As a testbed for our cooperative HOI framework, we design a cooperative carrying task that requires physically grounded coordination among multiple agents. As illustrated in Figure~\ref{fig:teaser}, multiple humanoid agents must jointly interact with and transport a large object, specifically a table of varying geometric shapes (e.g., round, rectangular, square). The task has sequential stages that collectively test agents cooperation: 1). \emph{Coordinated formation}: Agents begin at randomized initial locations and must autonomously navigate toward the table. Unlike CooHOI, which assumes oracle-provided per-agent hand target locations, our setup provides no explicit coordination assignment. Instead, 64 candidate contact points are uniformly sampled along the tabletop perimeter on its lower edge, from which agents must infer suitable positions among themselves to ensure stability during lifting. 2). \emph{Cooperative transport}: the agents collectively carry the table toward a designated goal location and then putting it down.

To accomplish this task, our overall reward formulation consists of several components: walking to object, contact, lifting, transport, and put-down—whose detailed expressions are provided in the supplementary. 

\subsubsection{Formation Reward}

Central to the success of this task is how the agents coordinate among themselves to walk into the object while spreading into a formation that promotes stable lifting.

\boldparagraph{Angular spread reward}
To facilitate this, we introduce an \emph{angular spread reward}. It provides a continuous learning signal that promotes the agents to evenly spread themselves around the table, and thus providing a stable support. For each agent, we find its nearest left and right agents and compute the 2D angular gaps to those neighbors about the table’s center, denoted as $\Delta\phi_i^{\text{ccw}}$ and $\Delta\phi_i^{\text{cw}}$.
For $m$ cooperating agents, the ideal spacing is $2\pi/m$. The reward is then formulated as:
\begin{equation}
r_{\text{ang}} =
\exp\!\Big(
    -k_\theta\, \tfrac{1}{2}\big[
        (\Delta\phi_i^{\text{ccw}} - \tfrac{2\pi}{m})^2
        + (\Delta\phi_i^{\text{cw}} - \tfrac{2\pi}{m})^2
    \big]
\Big).
\end{equation}

\begin{figure}[t!]
  \centering
   \includegraphics[width=1\linewidth]{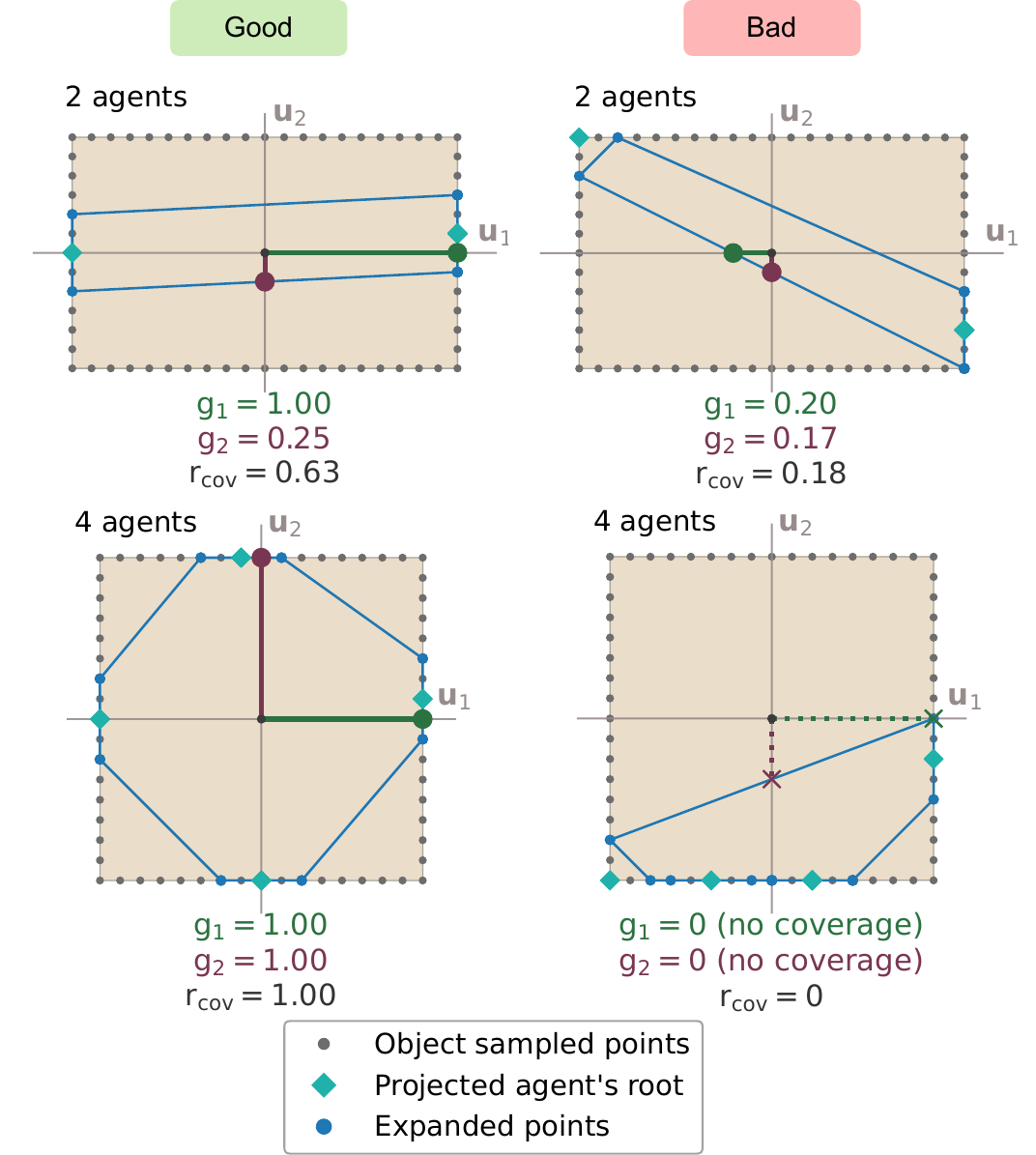}
   \caption{Illustration of our principal-axes coverage reward.}
   \label{fig:reward}
   \vspace{-0.3cm}
\end{figure}

\boldparagraph{Principal-axes coverage reward}
Humans most naturally walk forward or backward with symmetric gait, or move sideways through coordinated lateral steps, preferring movement aligned with their local body axes rather than diagonal directions. In cooperative transport, this tendency translates into agents positioning themselves into formations that maximize support along the object’s principal axes (its natural axes of rotational stability), and walking along those directions.

The angular spread reward, while effective in promoting balanced coverage for stable lifting, does not enforce these formations. We therefore introduce an additional reward that measures how well the agents’ support region spans the object’s principal axes, as illustrated in Figure~\ref{fig:reward}. First, each agent’s root position is projected to the nearest sampled points along the object's perimeter. This set of projected points defines a support polygon via a convex hull. To avoid degenerate case on two agents that can only form a line, each projected point is replaced by two points on its right and left along the perimeter (indices ±2). Let \(c\) denote the table’s center of mass in the \(x,y\) plane. We first compute the principal axes \(\mathbf{u}_1,\mathbf{u}_2\) centered on $c$ and measure the distances from \(c\) to the support polygon along both directions of each axis, \((d_i^+, d_i^-)\). When the support polygon does not extend across the table center of mass (lying entirely on one side), the distance becomes negative, indicating unbalanced or outside coverage. We therefore clip them as \(\tilde{d}_i^\pm=\max(0,d_i^\pm)\). With \((\ell_i^+,\ell_i^-)\) denoting the distances from the table center to the table boundary along the same axes, the per-axis coverage is:
\begin{equation}
g_i = \min\!\left(\frac{\tilde{d}_i^+}{\ell_i^+},\, \frac{\tilde{d}_i^-}{\ell_i^-}\right), \quad i\!\in\!\{1,2\},
\end{equation} 
and the resulting reward is $r_{\text{cov}} = \tfrac{1}{2}(g_1 + g_2)$. In the supplementary material, we provide a generalized formulation of $r_{\text{cov}}$ that supports irregular geometries and non-uniform mass distributions. 

$r_{\text{cov}}$ and $r_{\text{ang}}$ are designed to be complementary. While $r_{\text{cov}}$ remains valid under irregular geometries and mass distributions, it can be sparse when agents 
cluster on one side of the object (yielding zero reward). In contrast, $r_{\text{ang}}$ provides a continuous signal to encourage early dispersion and enable $r_{\text{cov}}$ to become non-zero. The final formation reward combines both effects with higher emphasis on $r_{\text{cov}}$:
\vspace{-0.2cm}
\begin{equation}
r_{\text{form}} = 0.25\, r_{\text{ang}} + 0.75\, r_{\text{cov}}.
\end{equation}

\section{Experiment}

\subsection{Implementation Details}

Here, we describe the key implementation details of our framework, including the observation states, architecture, and dataset tailored for the cooperative carrying task described in Section~\ref{sec:cooptask}. More implementation details are provided in the supplementary material.

\boldparagraph{Observation states} Every observing agent receives the following observation components, where each component is expressed in the agent's local coordinate frame:

\begin{itemize}

    \item \textbf{Self-proprioception} \(\in \mathbb{R}^{223}\): observing agent’s joint states and root kinematics as in standard AMP.

    \item \textbf{Object center} \(\in \mathbb{R}^{3}\): 3D location of the table center.
    
    \item \textbf{Candidate contact points} \(\in \mathbb{R}^{64\times3}\): 64 uniformly sampled points along the table perimeter on its lower edge. The first index is the point nearest to the agent’s root and the remainder are ordered counterclockwise.
    
    \item \textbf{Nearest hand-to-object points} \(\in \mathbb{R}^{2\times3}\): the candidate contact points nearest to each of the agent’s two hands.
    
    \item \textbf{Target object location} \(\in \mathbb{R}^{3}\): \(x,y\) goal of the table center and either the target height \(z\) or a binary indicator for lifting vs.\ putting-down.
    
    \item \textbf{Teammate cues} \(\in \mathbb{R}^{(n-1)\times9}\): for each of the \(n-1\) teammates, includes the \(x,y\) root position \((\mathbb{R}^{2})\), heading direction (6D rotation representation, \(\mathbb{R}^{6}\)), and relative angle between the observing agent’s root and the teammate’s root around the table center in the horizontal plane \((\mathbb{R}^{1})\). These cues are further encoded as teammate tokens.

\end{itemize}

\boldparagraph{Architecture}
We use the same observation states and transformer backbone for both the policy and critic networks, differing only in their final outputs. Each observation component is first encoded into a 64-dimensional token using separate three-layer MLP tokenizers with hidden sizes $[256, 128, 64]$. The input tokens include the 64-dimensional learnable embedding $e$, self-proprioception token, object token (combining object center, candidate contact points, and nearest hand-object points), target-location token, and a variable set of teammate tokens.

The transformer comprises three stacks of alternating self-attention and cross-attention layers, each with two attention heads and a 512-dimensional feed-forward block. The updated embedding $e$ is passed through an MLP with hidden sizes $[1024, 512, 28]$ to predict target joint rotations for PD control in the policy network, and $[1024, 512, 1]$ for critic. Both style discriminators, $D_\text{full}$ and $D_\text{mask}$ are implemented as MLPs with hidden sizes $[1024, 512, 1]$ following the standard AMP.

\boldparagraph{Humanoid and object models}
We adopt the Mujoco humanoid model with simplified ball hands without fingers. We design URDF models for three table geometries (square, rectangular, and round) used throughout the cooperative carrying experiments. The table mass ranges from 50 to 70~kg depending on the shape. The square table measures \(1.60~\text{m} \times 1.60~\text{m}\), the rectangular table \(2.00~\text{m} \times 1.20~\text{m}\), and the round table has a diameter of \(2.00~\text{m}\). The tabletop height is fixed at 0.82~m, slightly below the humanoid’s hand position in the default standing pose. This configuration allows our experiment to focus on multi-agent coordination rather than intricate contact interactions.

\begin{table*}[ht!]
\centering
\scriptsize 
\setlength{\tabcolsep}{4pt} 
\renewcommand{\arraystretch}{1.05}
\label{tab:coop-results}

\caption{
\textbf{Quantitative comparison across team sizes (2A, 4A, 8A).}
Our method achieves consistently high success rates, collective cooperation, and motion smoothness across all settings using a single unified policy. Unlike CooHOI* baselines, where agent formations are pre-defined, our agents must infer cooperation to establish stable formations autonomously, making the coordination requirement more demanding. Under the heavy-load setting (5× table weights), only our method demonstrates effective cooperation among eight agents. All results are averaged over 10,000 simulation episodes.
}
\label{tab:quantitative}
\vspace{-0.2cm}

\resizebox{\textwidth}{!}{%
\begin{tabular}{llccccccccccccc}
\toprule
\multirow{2}{*}{\textbf{Model}} &
\multirow{2}{*}{\begin{tabular}[l]{@{}l@{}}\textbf{Agents}\\\textbf{Formation}\end{tabular}} &
\multicolumn{3}{c}{\textbf{SR (\%)$\uparrow$ / $d$ ($\text{m}$)$\downarrow$}} &
\multicolumn{3}{c}{\textbf{$t_\text{coop}$ (\%)$\uparrow$}} &
\multicolumn{3}{c}{\textbf{$|J|$ ($\text{m/s}^3$) $\downarrow$}} &
\multicolumn{2}{c}{\textbf{SR$_{5\times}$(\%)$\uparrow$ / $d_{5\times}$ ($\text{m}$)$\downarrow$}} \\  
\cmidrule(lr){3-5} \cmidrule(lr){6-8} \cmidrule(lr){9-11} \cmidrule(lr){12-13}
 & & 2A & 4A & 8A & 2A & 4A & 8A & 2A & 4A & 8A & 4A & 8A \\
\midrule
CooHOI*-2 & Pre-defined &
97.5 / 0.19 & 73.2 / 1.49 & 10.1 / 4.25 &
90.3 & 54.6 & 1.0 &
48.3 & 85.0 & 189.9 &
0.0 / 6.38 & 0.4 / 6.00 \\
CooHOI*-4 & Pre-defined &
95.5 / 0.18 & 94.5 / 0.36 & 61.5 / 1.83 &
\textbf{96.0} & 92.1 & 27.2 &
40.7 & 38.6 & 96.7 &
1.2 / 5.34 & 14.2 / 4.26 \\
CooHOI*-8 & Pre-defined &
29.4 / 3.60 & 52.4 / 2.68 & 42.2 / 3.83 &
93.8 & 93.6 & 81.6 &
\textbf{39.5} & \textbf{36.5} & 45.2 &
0.1 / 6.22 & 4.1 / 5.73 \\
\midrule
\textbf{Ours} & Learned coop. &
\textbf{99.1} / \textbf{0.06} & \textbf{99.2} / \textbf{0.08} & \textbf{97.5} / \textbf{0.18} &
95.2 & \textbf{96.1} & \textbf{90.1} &
51.0 & 44.7 & \textbf{34.2} &
\textbf{3.5} / \textbf{4.78} & \textbf{81.1} / \textbf{0.49} \\
\bottomrule
\end{tabular}%
}

\end{table*}

\begin{figure*}[ht!]
  \centering
   \includegraphics[width=0.98\linewidth]{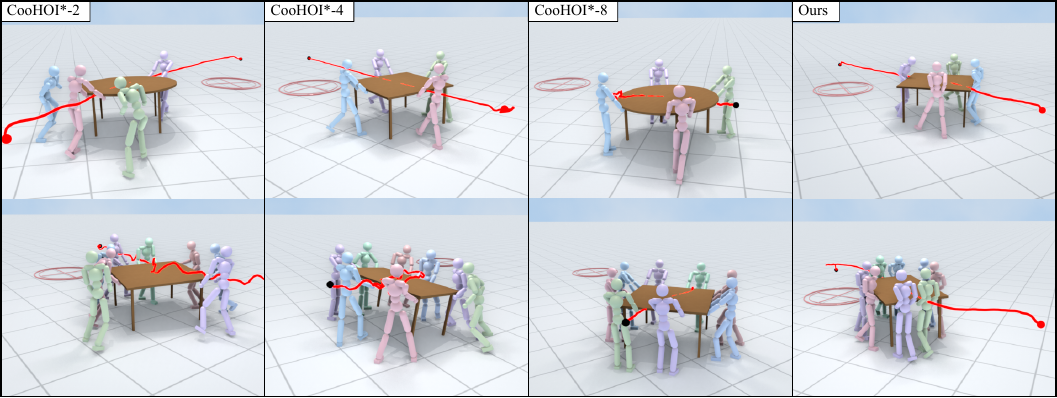}
   \caption{
\textbf{Qualitative comparison across 4-agent (top) and 8-agent (bottom) configurations.} Our method produces synchronized and stable teamwork across both cases, whereas the CooHOI* baselines exhibit limited or ineffective cooperation. Red line indicates the table’s movement trajectory, and the black dot marks its final position at the end of each episode.} 
   \label{fig:qualitative}
   \vspace{-0.2cm}
\end{figure*}

\boldparagraph{Reference motions}
Our reference motions are from AMASS dataset~\cite{mahmood2019amass}. Following CooHOI, we adopt 9 walking-related motions from the ACCAD subset and their temporally reversed versions for backward walking, along with 3 sideways-walking motions from the CMU subset. To enable lowering the upper body toward the table and lifting back up, we use 3 pickup motions from the ACCAD subset. These sequences are trimmed before reaching excessively low postures and then reversed to generate the corresponding lifting motions.

\subsection{Evaluation}

\boldparagraph{Simulation setup} We train a unified policy for 2-8 agents using our \OURS{} framework. At the start of each episode, agents are randomly initialized on a circle of radius $8$ m centered on the object, and the target location is placed in a random direction $[3,10]$ m away from the table center. Each evaluation episode runs for 600 simulation timesteps.

\boldparagraph{Baselines} To obtain non-trivial baselines, we substantially adapt the CooHOI~\cite{gao2024coohoi} framework, denoted as CooHOI*. First, we incorporate our masked AMP strategy to enable diverse realistic locomotion skills for the multi-agent table-carrying. Second, we design a specialized reward function so that any agent, from any initial position, can reach a specified contact point without colliding the table.

Training follows a two-stage pipeline. In the first stage, a single agent is trained to acquire foundational skills: approaching a given contact point, lifting the table, and pushing or dragging it toward the goal. In the second stage, the same policy is extended to multi-agent settings, where coordination emerges implicitly through object dynamics.  We train three variants, denoted as CooHOI*-$n$, where $n \in \{2, 4, 8\}$ indicates the number of cooperating agents. In each configuration, agents are assigned fixed contact points that are manually selected to provide maximum coverage along the object’s principal axes. Note that this setup eliminates the need for coordinated formation in the original task, as agents are guided by oracle-defined contact points. Thus, we do not enforce formation reward for CooHOI*, while other rewards are kept the same as ours. Further details of CooHOI* are in the supplementary material.


\boldparagraph{Metrics}
We evaluate the cooperative carrying task using four quantitative metrics that capture task success, cooperation quality, and motion smoothness:

\begin{itemize}

    \item \textbf{Success rate (SR):} Fraction of episodes with successful task completion, where the object-target distance reaches $0.03~\text{m}$. Higher is better.
    
    \item \textbf{Distance to target ($d$):} Euclidean distance (in meters) between the table center and target location at the end of episode. For successful episodes, $d$ is reported as $0.03$ since rewards are not enforced after putdown and baseline models exhibit erratic movements. Lower is better.

    \item \textbf{Cooperative time ratio ($t_\text{coop}$):} Fraction of the transport duration during which all agents maintain contact with any of the 64 contact points, indicating consistent collective contribution. Higher is better.

    \item \textbf{Mean absolute jerk ($|J|$):} Average magnitude of the third derivative of the 64 contact point trajectories, capturing transport motion smoothness. Lower is better.
\end{itemize}

\boldparagraph{Results} We test each model on the cooperative carrying task with 2, 4 and 8 cooperating agents. Table~\ref{tab:quantitative} presents the quantitative results averaged over 10,000 simulations for each cooperative scenario. Our method consistently achieves high success rates, collective cooperation, and smooth motion across all configurations, demonstrating the effectiveness of a single unified policy obtained from our framework. In contrast, the CooHOI* baselines exhibit strong dependence on the specific team size they are trained for. CooHOI*-2 performs well only for two agents but fails to exhibit coordinated behaviors when scaled to larger teams. Similarly, CooHOI*-4 maintains good performance up to four agents but deteriorates sharply beyond that configuration, while CooHOI*-8 struggles to establish effective coordination even within its own setup. We further evaluate the models under a heavy-load setting (5× table weight). The task becomes too challenging for smaller teams, which are barely capable of lifting the table. Notably, only our method demonstrates meaningful cooperation among eight agents, which collectively handle the increased load and achieve a high success rate.

The qualitative results in Figure~\ref{fig:qualitative} further highlight the behavioral differences between our approach and CooHOI* baselines. CooHOI*-2 exhibits competing behaviors between agent pairs, where each pair attempts to move the table independently, resulting in uneven motion and frequent loss of contact. CooHOI*-4 fails to exhibit synchronized cooperation in larger teams, leading to unstable movements. CooHOI*-8 struggles to coordinate effectively when moving the table toward the target, often generating conflicting forces among agents. In contrast, our method achieves globally coherent motion: all agents lift, stabilize, and transport the object as a unified team. These results demonstrate that scalable cooperation across varying team sizes and coordination demands is achieved with the unified decentralized policy trained in our framework. In the supplementary material, we also present more comprehensive experimental results, including robustness to unseen setups and zero-shot generalization to 16-agent configuration.

\subsection{Ablation Study}

To better understand the contributions of the masked AMP strategy and the proposed formation reward, we conduct an ablation study to examine the effect of each component.

\begin{figure}[h!]
    \centering
    \includegraphics[width=0.8\linewidth]{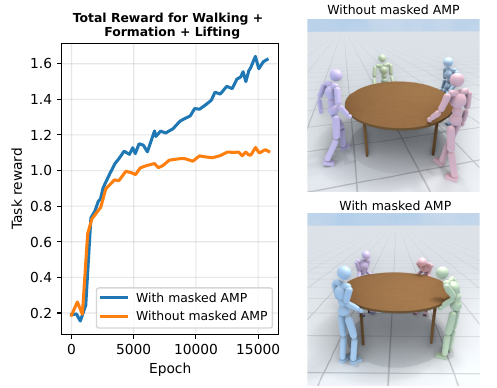}
    \caption{
    \textbf{Ablation on the masked AMP strategy.} Comparison between models trained with and without masked AMP, showing improved task rewards and successful hand-object interactions when masking is applied.
    }
    \label{fig:plot_ablation}
\end{figure}

\boldparagraph{Masked AMP}
As shown in Figure~\ref{fig:plot_ablation}, incorporating masked AMP significantly improves the overall success rate of the lifting stage. By masking object-interacting body parts during discriminator updates, the policy learns to control hand-object interactions through task rewards rather than being limited by the over-constrained single-human full-body reference motions. Without masking, conflicting objectives between motion realism and object interaction often lead to failed coordination during lifting. Masked AMP alleviates this issue by enabling greater diversity in hand-object interactions. In the subsequent stages, it allows the policy to learn more varied coordination patterns, such as walking or stepping in different directions while carrying the table, despite relying only on single-agent reference motions.

\begin{figure}[h!]
    \centering
    \includegraphics[width=1\linewidth]{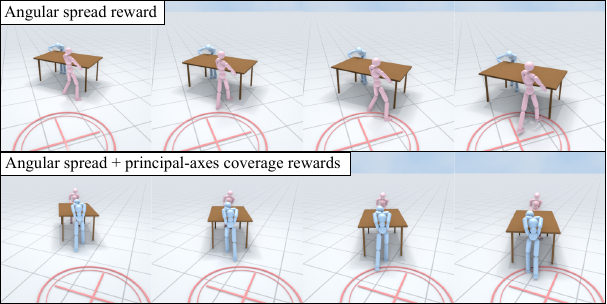}
    \caption{
    \textbf{Formation reward comparison.} Adding principal-axes coverage reward produces stable formations aligned with the object’s principal axes, facilitating learned natural locomotion.}
    \label{fig:ablation_formation}
\end{figure}

\boldparagraph{Formation Reward}
Figure~\ref{fig:ablation_formation} compares policies trained with and without the proposed principal-axes coverage reward. When the policy is only trained with angular spread reward, agents do not always distribute themselves along the object’s principal axes. During mid-training, this setup often produces object trajectories with excessive rotation around the vertical axis. The policy eventually stabilizes but converges to an unnatural diagonal stepping patterns misaligned with the object’s principal axes. These unnatural stepping patterns persist across different scenarios and team sizes.

In contrast, incorporating the principal-axes coverage reward encourages agents to align their formations along the object’s natural axes of rotational stability. As a result, the team learns to walk in coordinated directions with natural symmetric gaits and balanced support.

\section{Conclusion}
We present \OURS{}, a unified framework for scalable cooperative human-object interaction that enables a single decentralized policy to generalize across varying team sizes and object configurations. By introducing a Transformer-based policy architecture with teammate tokens, our approach efficiently incorporates teammate cues in a scalable manner, enabling effective inter-agent coordination across varying team sizes. The masked AMP strategy broadens the motion diversity achievable from single-human reference data, while the principal-axes coverage reward encourages stable and natural formations during cooperative transport. Through comprehensive experiments in cooperative carrying task, we demonstrated that \OURS{} achieves coherent, stable, and diverse coordination behaviors across a wide range of multi-agent settings. We believe this work establishes a foundation for scalable, physics-based multi-humanoid control and opens new opportunities for both embodied intelligence and multi-character animation in virtual environments.

\noindent{\textbf{Acknowledgment.}} This research is supported by the National Research Foundation (NRF) Singapore, under its NRF-Investigatorship Programme (Award ID. NRF-NRFI09-0008), and the Tier 2 grant MOET2EP20124-0015 from the Singapore Ministry of Education.


{
    \small
    \bibliographystyle{ieeenat_fullname}
    \bibliography{main}
}


\maketitlesupplementary

\section{Training with Various Team Sizes}

\subsection{Team-Size Advantage Normalization}

PPO~\cite{schulman2017proximal} algorithm computes the advantage term $A_t$, which measures how much better an action performs relative to the expected return of the policy’s actions in the same state, as estimated by the critic network. The advantages are computed from trajectories collected over a finite time horizon, which determines how far into the future rewards are accumulated. Because their scale can vary across trajectories, the advantages are typically normalized across a batch of trajectories to stabilize training, given as:
\[
A_t \leftarrow \frac{A_t - \mu(A)}{\sigma(A) + \epsilon},
\]
where $\mu(A)$ and $\sigma(A)$ are the mean and standard deviation computed over the batch. 

\begin{figure}[h!]
    \centering
    \includegraphics[width=0.65\linewidth]{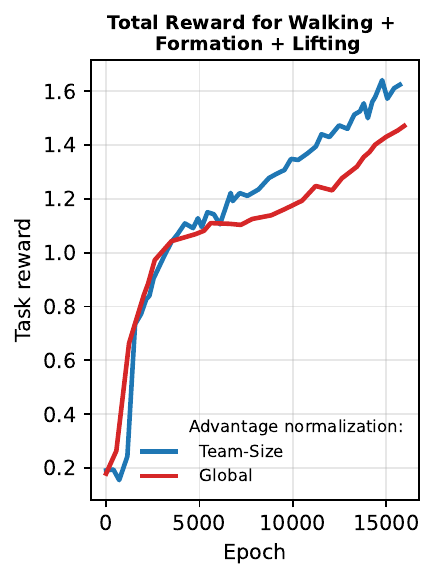}\vspace{-0.2cm}
    \caption{
    Comparison of task reward curves for models trained with team-size and global advantage normalizations.
    }
    \vspace{-0.2cm}
    \label{fig:plot_norm}
\end{figure}

In our framework, training batches can include data from teams of different sizes, each producing rewards with distinct scales and variances. Normalizing all advantages together across such heterogeneous data can distort their relative magnitudes and influence the accuracy of the policy update signal. Thus, we normalize advantages separately for each team size $n$:
\[
A_t^{(n)} \leftarrow 
\frac{A_t^{(n)} - \mu_n(A)}{\sigma_n(A) + \epsilon}.
\] As seen in Figure~\ref{fig:plot_norm}, the team-size advantage normalization results in higher task reward.

\subsection{Environment Instantiation}

\begin{figure}[h!]
  \centering
   \includegraphics[width=1\linewidth]{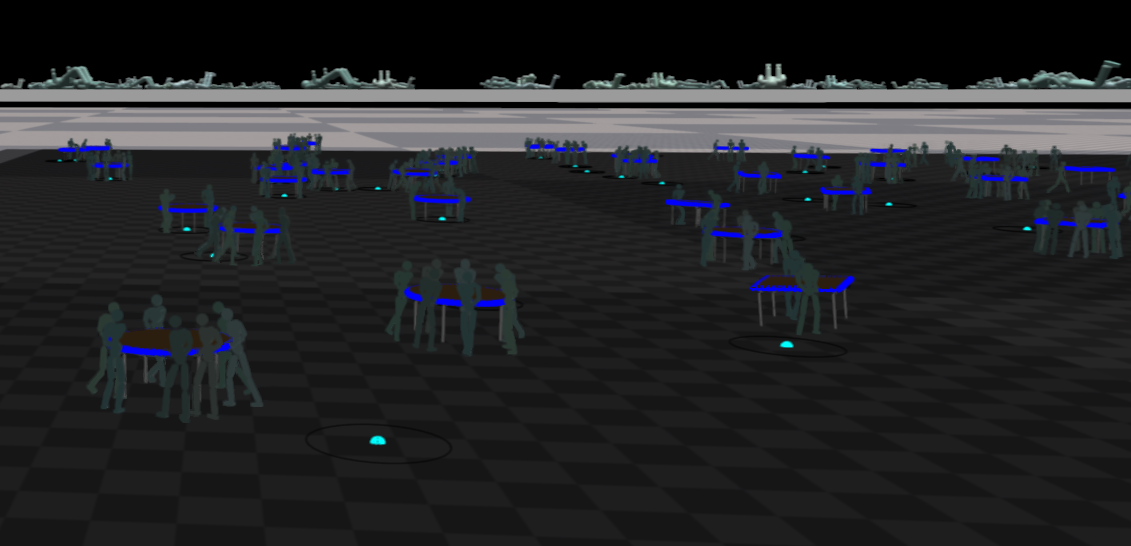}
   \caption{Environment setup in IsaacGym using a dummy ceiling to support flexible team-size training. Extra agents are moved to the ceiling and excluded from observations, rewards, and gradient updates.}
   \label{fig:sim}
\end{figure}

We use IsaacGym~\cite{makoviychuk2021isaac} simulator to train our model. A current limitation of IsaacGym is that each environment in the parallel training must contain the same number of actors, including the humanoid agents and objects. To address this limitation and enable the any team-size unified policy training, we add a dummy ceiling plane in each environment and instantiate a fixed number of agents $N$. For any environment that requires a smaller team size $n$, we place the remaining $N-n$ agents on the dummy ceiling. These agents are ignored for the reward calculation, observation states, and gradient computation. Additionally, their PD controllers are disabled. See Figure~\ref{fig:sim} for an illustration.

\section{Reward Functions}

Here, we detail all reward components used in the cooperative carrying task, excluding the formation reward $r_\text{form}$, angular spread reward $r_\text{ang}$, and principal-axes coverage reward $r_\text{cov}$ already described in the main paper.

\subsection{Walking Toward Object}

After initialization, each humanoid starts at some distance from the table and is encouraged to walk to the object before the lifting phase. The walking reward is decomposed into three terms that shape the position, velocity, and facing of each agent.

\boldparagraph{Position}
For each agent, we locate the nearest sampled point along the table perimeter, denoted as $\mathbf{p^*}$, and compute the distance to the agent's root $\mathbf{x_{\text{root}}}$ in $x\text{-}y$ plane, $d = \lVert \mathbf{x_{\text{root}}} - \mathbf{p^*} \rVert_2$. The agent is encouraged to stand at a target gap $d_{\text{gap}}=0.3\text{ m}$ from $\mathbf{p^*}$ by penalizing the squared deviation $ \Delta_{\text{gap}} = (d - d_{\text{gap}})^2$. The position reward is:
\begin{equation}
r_{\text{walk}}^{\text{pos}} =
\begin{cases}
\exp\bigl(-2.0 \Delta_{\text{gap}}\bigr), & \Delta_{\text{gap}} > 0.04 \text{ m},\\[4pt]
1, & \Delta_{\text{gap}} \le  0.04 \text{ m}.
\end{cases}
\end{equation}
This term acts as an attractive potential that pulls each agent toward the table. It must be balanced by the formation reward $r_{\text{form}}$ to ensure that agents spread out while still converging to their appropriate standing regions before lifting.

\boldparagraph{Velocity}
Let $\mathbf{v} \in \mathbb{R}^2$ be the $x,y$ root velocity. We define a desired walking direction $\mathbf{u^*} \in \mathbb{R}^2$ as the inward unit normal in $x\text{-}y$ plane associated with the nearest perimeter point $\mathbf{p^*}$. The agent’s directional speed $s$ is computed by projecting $\mathbf{v}$ onto $\mathbf{u^*}$,
$s = \mathbf{u^{*\top}} \mathbf{v}$.
We then encourage the agent to move toward the table within a preferred speed range from $s^*_\text{low} = 1.5 \text{ m/s}$ to $s^*_\text{high} = 2.5 \text{ m/s}$. The deviation from this range is expressed using ReLU functions, $\delta_\text{vel}= \max(0, s^*_\text{low} - s) + \max(0, s - s^*_\text{high}).$
The velocity reward is:
\begin{equation}
r_{\text{walk}}^{\text{vel}} =
\begin{cases}
0, & s \le 0,\\[4pt]
1, & \Delta_{\text{gap}} \le 0.04 \text{ m},\\[4pt]
 \exp\bigl(-2.0 \, \delta_\text{vel}^2\bigr), & \text{otherwise}.
\end{cases}
\end{equation}

\boldparagraph{Facing}
We compute the agent’s facing direction by extracting the heading component of its root orientation. Let $\mathbf{f} \in \mathbb{R}^2$ be the facing direction in the $x\text{-}y$ plane. We define two target facing directions: the inward normal $\mathbf{u^*}$ at $\mathbf{p^*}$ for near-view alignment, and the direction from the agent toward the table center, $
\mathbf{c^*} = \frac{\mathbf{p_{\text{center}}} - \mathbf{x_{\text{root}}}}{\lVert \mathbf{p_{\text{center}}} - \mathbf{x_{\text{root}}} \rVert_2}$. The facing reward is computed as:
\begin{equation}
r_{\text{walk}}^{\text{face}} =
\begin{cases}
 \max(0, \mathbf{u^{*\top}}\mathbf{f}) , & d \le 1.0 \text{ m},\\[4pt]
\max(0, \mathbf{c^{*\top}} \mathbf{f}), & d > 1.0 \text{ m}.
\end{cases}
\end{equation}

\subsection{Hand Contact Preparation}

After the agents are close to the table with $\lVert\mathbf{x_{\text{root}}} - \mathbf{p^*} \rVert_2 \le 1.0 \text{ m} $, agents are encouraged to reach the hands towards to the table contact points and maintain a reasonable hand configuration for lifting.

\boldparagraph{Hand reaching}
For each agent, let $\mathbf{h}_{j} \in \mathbb{R}^3$, $j \in \{\text{L}, \text{R}\}$, be the left and right hand positions, and $\{\mathbf{q}_k\}_{k=1}^{64}$ the 64 candidate contact points. We first find the nearest contact point for each hand and its distance,
$d^{\text{hand}}_{j} = \min_k \lVert \mathbf{h}_{j} - \mathbf{q}_k \rVert_2$.
A proximity term encourages both hands to approach the contact point:
\begin{equation}
 r_{\text{prox}} = \frac{1}{2} \sum_{j} \exp(-5.0 d^{\text{hand}}_{j}).   
\end{equation}

In addition, we encourage the hands to reach the lower edge of the table rather than drifting onto the tabletop surface. For each hand $j \in \{\text{L}, \text{R}\}$ with position $\mathbf{h}_j \in \mathbb{R}^3$, let $\mathbf{p}^*_j \in \mathbb{R}^3$ be its nearest sampled perimeter point on the table. We define a contact direction $
\hat{\mathbf{v}}_j = \frac{\mathbf{h}_j - \mathbf{p}^*_j}{\lVert \mathbf{h}_j - \mathbf{p}^*_j \rVert_2}$. Let $\mathbf{e}_z = (0, 0, 1)^\top$ be the world-up direction. We compute $\cos\theta_j = \hat{\mathbf{v}}_j^\top \mathbf{e}_z$,
which measures how much the hand moves upward relative to its associated contact point. The per-hand vertical alignment score is then defined as:
\begin{equation}
r_{\text{above}, j} =
\begin{cases}
\exp\bigl(-3.0 \cos\theta_j\bigr), & \cos\theta_j > 0,\\[4pt]
1, & \cos\theta_j \le 0.
\end{cases}
\end{equation}
The combined term over both hands is:
\begin{equation}
r_{\text{above}} = \frac{1}{2}\bigl(r_{\text{above}, \text{L}} + r_{\text{above}, \text{R}}\bigr).
\end{equation}

\boldparagraph{Hand separation} We also encourage a target horizontal separation between the two hands. First, we compute the horizontal separation in the $x\text{-}y$ plane, $d_{\text{hand}} = \lVert (\mathbf{h}_{\text{L}} - \mathbf{h}_{\text{R}})_{xy} \rVert_2$.
We encourage the hands to remain within a preferred separation interval 
$d^*_{\text{low}} = 0.4\text{ m}$ and $d^*_{\text{high}} = 0.6\text{ m}$. Deviations from this interval are expressed using ReLU functions, $\delta_{\text{sep}} = \max(0, d^*_{\text{low}} - d_{\text{hand}}) + \max(0, d_{\text{hand}} - d^*_{\text{high}})$. We then obtain a hand separation reward:
\begin{equation}
r_{\text{sep}} =
\exp\bigl(-5.0 \delta_{\text{sep}}^2\bigr).
\end{equation}
To encourage consistent lifting, we penalize vertical mismatch between the two hands. Let $z_{\text{L}}$ and $z_{\text{R}}$ be their heights, and the reward is defined as:
\begin{equation}
r_{\text{same-z}} = \exp\bigl(-20.0 (z_{\text{L}} - z_{\text{R}})^2\bigr).
\end{equation}

\boldparagraph{Combined reward}
The combined hand preparation reward is:
\begin{equation}
r_{\text{hand}} = r_{\text{prox}} \times r_{\text{above}} \times r_{\text{sep}} \times r_{\text{same-z}},
\end{equation}
which requires all four terms to be satisfied simultaneously.

\subsection{Contact and Lifting}

Once the hands are placed near the table edge, additional rewards are activated so that the agents establish contact and lift the table to a desired height.

\boldparagraph{Contact activation}
Let $d^{\text{hand}}_j$ be the nearest hand-to-contact distance defined earlier. A per-hand contact score is computed as $
\gamma_j = \max\Bigl(0, 1 - \frac{d^{\text{hand}}_j}{0.06 \text{ m}}\Bigr)$. We then define a contact reward as the minimum of the per-hand contact scores across the two hands:
\begin{equation}
r_{\text{contact}} = \min(\gamma_{\text{L}}, \gamma_{\text{R}}),
\end{equation}
and contact indicator for each hand:
\[
m_j =
\begin{cases}
1, & d^{\text{hand}}_j < 0.04 \text{ m},\\[4pt]
0, & \text{otherwise},
\end{cases}
\]
which is used to gate the subsequent lifting and transport rewards.

\boldparagraph{Lifting height}
After contact is established, the hands should lift the table to a target height. Let $\hat{z}_j$ be the height of the contact point associated with hand $j$, and the target lifting height $z^*_{\text{lift}} = 0.94\text{ m}$. We obtain a lifting reward for each hand:

\begin{equation}
\rho_j = \exp(-5.0 \, \lvert \hat{z}_j - z^*_{\text{lift}} \rvert).
\end{equation}
Only hands with valid contact contribute. Therefore, the combined lifting reward is given as:
\begin{equation}
r_{\text{lift}} = \frac{1}{2}\bigl(m_{\text{L}} \rho_{\text{L}} + m_{\text{R}} \rho_{\text{R}}\bigr).
\end{equation}

\subsection{Collective Transport}

\boldparagraph{Transport} Once all agents establish contact with the table using both hands, they are encouraged to move the object toward a target location collectively. Let $\mathbf{x}_{\text{obj}} \in \mathbb{R}^2$ be the $x,y$ table position and $\mathbf{x}_{\text{tar}} \in \mathbb{R}^2$ the target location. We define define the transport reward as:
\begin{equation}
r_{\text{transport}} =
\begin{cases}
\exp\Bigl(-0.15 \, \lVert \mathbf{x}_{\text{tar}} - \mathbf{x}_{\text{obj}} \rVert_2^2\Bigr),
&
\begin{array}{l}
m_{\text{L}} = m_{\text{R}} = 1 \\
\text{for all agents},
\end{array}\\[10pt]
0, & \text{otherwise}.
\end{cases}
\end{equation}

\boldparagraph{Carrying alignment}
While not strictly required for transport, we include a carrying alignment reward that encourages at least one agent to face toward the target direction while carrying. We identify this agent as the agent farthest from the target. Let $\mathbf{f} \in \mathbb{R}^2$ be this agent’s facing direction in the $x\text{-}y$ plane, and the desired transport direction:
\[
\mathbf{u}_{\text{tar}} =
\frac{\mathbf{x}_{\text{tar}} - \mathbf{x}_{\text{obj}}}
     {\lVert \mathbf{x}_{\text{tar}} - \mathbf{x}_{\text{obj}} \rVert_2}.
\]
We compute the alignment reward:
\begin{equation}
r_{\text{align}} =
\begin{cases}
\max\bigl(0,\, \mathbf{u}_{\text{tar}}^\top \mathbf{f}\bigr), &
\begin{array}{l}
m_{\text{L}} = m_{\text{R}} = 1 \text{ for all agents}\\
\text{and } \lVert \mathbf{x}_{\text{tar}} - \mathbf{x}_{\text{obj}} \rVert_2 \ge 0.5\text{ m},
\end{array}
\\[12pt]
1, &
\begin{array}{l}
m_{\text{L}} = m_{\text{R}} = 1 \text{ for all agents}\\
\text{and } \lVert \mathbf{x}_{\text{tar}} - \mathbf{x}_{\text{obj}} \rVert_2 < 0.5\text{ m},
\end{array}
\\[12pt]
0, & \text{otherwise}.
\end{cases}
\end{equation}

Both $r_{\text{transport}}$ and $r_{\text{align}}$ are shared across all agents. During transport, when $m_{\text{L}} = m_{\text{R}} = 1 \text{ for all agents}$, we set $r^{\text{face}}_\text{walk} = 1.0$ so that agents can adjust flexible heading directions while carrying the object collectively.

\subsection{Putdown}

Once the object reaches the target location, agents must putdown the table and release their hands from the table. Thus, we introduce putdown reward once the object reaches target: $
\lVert \mathbf{x}_{\text{tar}} - \mathbf{x}_{\text{obj}} \rVert_2 < 0.03\text{ m}$.

\boldparagraph{Hand release}
Let $z_{j}$ be the height of hand $j \in \{\mathrm{L},\mathrm{R}\}$ and 
$z^*_{\text{put}} = 0.65\text{ m}$ the target hand height during putdown.
Let $d^{\text{hand}}_{j}$ be the nearest hand–table distance defined earlier. We compute the hand-release reward as:
\begin{equation}
r_{\text{put}}^{\text{release}} =
\begin{cases}
1, &
\begin{array}{l}
\hspace{-4.0em} d^{\text{hand}}_{\mathrm{L}} > 0.07\text{ m} \\[2pt]
\hspace{-4.0em} \text{and } d^{\text{hand}}_{\mathrm{R}} > 0.07\text{ m},
\end{array}
\\[12pt]
\displaystyle \min_{j \in \{\mathrm{L},\mathrm{R}\}}
\exp\!\bigl(-5.0 \, |z_j - z^*_{\text{put}}|\bigr),
& \text{otherwise}.
\end{cases}
\end{equation}

\boldparagraph{Zero velocity}
During putdown, we also encourage agents to stop moving by applying the following reward:  
\begin{equation}
r_{\text{put}}^{\text{vel}} = \exp\bigl(-2\, \lVert \mathbf{v} \rVert_2\bigr),
\end{equation}
where $\mathbf{v}$ is the agent's $x,y$ root velocity.

\boldparagraph{Combined reward}
The final putdown reward is a weighted combination of the hand–release and zero–velocity terms:
\begin{equation}
r_{\text{put}} =
0.8\, r_{\text{put}}^{\text{release}}
\;+\;
0.2\, r_{\text{put}}^{\text{vel}}.
\end{equation}

\subsection{Total Task Reward}

The task reward for the cooperative carrying task is aggregated as follows:

\begin{equation}
\begin{aligned}
r^{\text{task}} =\;&
0.2\, r^{\text{pos}}_{\text{walk}}
+ 0.4\, r^{\text{vel}}_{\text{walk}}
+ 0.2\, \sqrt{\,r^{\text{face}}_{\text{walk}}\, r_{\text{ang}}\,}
+ 0.6\, r_{\text{form}} \\
&+ 0.7\, (r_{\text{hand}} r_{\text{cov}})
+ 0.7\, r_{\text{contact}}
+ 0.7\, (r_{\text{lift}} r_{\text{cov}}) \\
&+ 1.0\, r_{\text{transport}}
+ 0.4\, r_{\text{align}}
+ 1.0\, r_{\text{put}}.
\end{aligned}\label{eq:task}
\end{equation}

\section{Generalized Principal-Axes Coverage Reward}
\label{sec:supp_principal_axes_cov}

We elaborate the components to obtain the generalized principal-axes coverage reward $r_{\text{cov}}$ that supports irregular geometries (including concave shapes such as L-shape) and non-uniform mass distributions. 

\boldparagraph{Center of mass} Let $\mathcal{X} = \{\mathbf{x}_k \in \mathbb{R}^2\}_{k=1}^N$ denote a set of 2D points sampled from the object’s $x$–$y$ plane (e.g., the tabletop surface). Each point may optionally carry a mass weight $w_k > 0$ representing local density. Uniform density corresponds to $w_k = 1$. The planar center of mass is obtained as $
\mathbf{c}
=
\frac{\sum_{k=1}^N w_k \mathbf{x}_k}
     {\sum_{k=1}^N w_k}.
$

\boldparagraph{Principal-axes}
Next, we obtain the principal-axes $\mathbf{u}_1$ and $\mathbf{u}_2$ from the eigenvectors of the real and symmetric object's planar inertia matrix {\scriptsize
$\mathbf{I} = \begin{bmatrix} I_{xx} & I_{xy} \\ I_{xy} & I_{yy} \end{bmatrix}$}. Let $\tilde{\mathbf{x}}_k = \mathbf{x}_k - \mathbf{c}$ denote the centered coordinates with components $(\tilde{x}_k, \tilde{y}_k)$. The inertia components are computed as
$I_{xx} = \sum_k w_k \tilde{y}_k^2$,
$I_{yy} = \sum_k w_k \tilde{x}_k^2$,
and
$I_{xy} = - \sum_k w_k \tilde{x}_k \tilde{y}_k$. We then compute the eigen decomposition of $\mathbf{I}$ and define $\mathbf{u}_1$ as the eigenvector associated with the smallest eigenvalue, and $\mathbf{u}_2$ as the remaining orthonormal eigenvector.

\boldparagraph{Boundary extents}
We compute the boundary extents $\ell_i^{+}$ and $\ell_i^{-}$ along each principal axis $\mathbf{u}_i$ in a manner that remains well-defined for irregular and concave geometries. To this end, we compute the convex hull $\mathcal{H}$ of the object boundary in the planar domain. The boundary extents are the maximum distances from the center of mass $\mathbf{c}$ to the convex hull $\mathcal{H}$ along the positive and negative directions of $\mathbf{u}_i$.

\section{Additional Implementation Details}

\subsection{Training Strategy}

Training the unified policy directly with up to eight agents is computationally inefficient due to the long-horizon nature of the cooperative carrying task. We therefore adopt a sequential training with different stages that progressively approaches task completion and increases team size, with early termination triggered whenever any agent falls or table topples. All stages below are trained using a single NVIDIA A100 GPU.

\boldparagraph{Stage 1} We first train environments instantiated with 1-4 agents to acquire core navigation, contact, and lifting behaviors. At this stage, the transport, alignment, and putdown rewards are disabled (i.e., $r_{\text{transport}} = r_{\text{align}} = r_{\text{put}} = 0$).  The full-body discriminator $D_{\text{full}}$ is supervised using only forward and sideways walking reference motions, while the masked discriminator $D_{\text{mask}}$ additionally receives pickup motions. The target-location token is also masked out during this stage. Training runs for approximately 1.5 days with an episode length of 400 timesteps.

\boldparagraph{Stage 1+2} Next, we continue training with 2-4 agents to proceed with the coordinated transport and putdown. We enable the remaining task components, including $r_{\text{transport}}$, $r_{\text{align}}$ and $r_{\text{put}}$, as well as unmasking the target-location token. Backward walking reference motions are added to supervise $D_{\text{mask}}$ to improve locomotion diversity while carrying the object. This stage converges in roughly 5 days with an episode length of 600 timesteps.

\boldparagraph{Fine-tuning with up to 8 agents} Finally, we fine-tune the unified policy in environments instantiated with 2-8 agents to refine coordination patterns and stabilize collective transport for larger teams. This fine-tuning stage takes about 3 days.

\subsection{Training hyperparameters}

We train all stages using 1024 parallel environments.
For PPO, the minibatch size is set to 16384 when training with up to four agents, and reduced to 8192 for training with up to eight agents.
For both PPO and AMP updates, observations belonging to deactivated agents (i.e., those placed above the ceiling) are excluded from the minibatches. For AMP, the reference-motion minibatch size is 4096, and the policy-observation minibatch is set to $1.5\times$ 4096, but excluding the deactivated agents.

Unless otherwise noted, all remaining hyperparameters follow CooHOI~\cite{gao2024coohoi}.
We list the key values in Table~\ref{tab:hyperparams} for completeness.

\begin{table}[h!]
\centering
\small
\setlength{\tabcolsep}{8pt}
\renewcommand{\arraystretch}{1.1}
\caption{
Key training hyperparameters in our experiment.
}
\label{tab:hyperparams}
\vspace{-0.1cm}
\begin{tabular}{ll}
\toprule
\textbf{Hyperparameter} & \textbf{Value} \\
\midrule
Horizon length             & 32 \\
Optimizer                  & Adam~\cite{kingma2015adam} \\
Learning rate              & $2\times 10^{-5}$ \\
Task reward weight         & 0.5 \\
Style reward weight        & 0.5 \\
PPO clip threshold $(\epsilon)$ & 0.2 \\
Discount factor $(\gamma)$   & 0.99 \\
GAE parameter $(\lambda)$    & 0.95 \\
\bottomrule
\end{tabular}
\end{table}

\section{CooHOI* Baseline}

\boldparagraph{Architecture}
CooHOI* follows the same Transformer-based backbone as our method for both the policy and the critic, but replaces the cross-attention with self-attention layer without incorporating teammate tokens. This design mimics the original CooHOI formulation where cooperation emerges solely from the shared dynamics of the object.

\boldparagraph{Approach-angle reward}
We design an approach-angle reward to guide each agent toward its designated contact point while avoiding collision with the table. Let $\mathbf{p}_\text{des} \in \mathbb{R}^2$ be the $x,y$ coordinate of the designated point. We first calculate the normalized $x,y$ direction from the object to agent's root: $\hat{\mathbf{a}}_{o} = 
\frac{\mathbf{x}_{\text{root}} - \mathbf{x}_{\text{obj}}}{\lVert \mathbf{x}_{\text{root}} - \mathbf{x}_{\text{obj}} \rVert_2}$, as well as the normalized $x,y$ direction the object to the designated point: $\hat{\mathbf{p}}_{o} = 
\frac{\mathbf{p}_{\text{des}} - \mathbf{x}_{\text{obj}}}{\lVert \mathbf{p}_{\text{des}} - \mathbf{x}_{\text{obj}}\rVert_2}$. We then calculate the approach-angle reward based on the cosine similarity between the two directions:

\begin{equation}
r_{\text{approach}}
= \frac{\hat{\mathbf{a}}_{o}^{\top} \hat{\mathbf{p}}_{o} + 1}{2}.
\end{equation} 
This yields $r_{\text{approach}} = 1$ when the agent is perfectly aligned toward its target contact point ($\theta = 0^\circ$) and $r_{\text{approach}} = 0$ when it faces the opposite direction ($\theta = 180^\circ$), effectively avoiding collision with the table  when agents are initialized in random positions.

The aggregated task reward follows the same structure as Equation~\ref{eq:task}, except that $r_{\text{form}}$, $r_{\text{ang}}$, and $r_{\text{cov}}$ are replaced by the approach-angle reward $r_{\text{approach}}$.

\boldparagraph{Training strategy} We follow a two-stage training procedure as in CooHOI. In the first stage, a single agent is trained to acquire foundational locomotion and manipulation skills, including approaching the table, maneuvering toward the designated contact point, establishing contact, lifting (or tilting) the table to the target height, and subsequently pushing or dragging it toward the goal.
To simplify learning, the friction between the table legs and the ground is set to zero and the table mass is reduced by half during this stage. Training runs for approximately 3 days. 

Multi-agent cooperation is then introduced in the second stage by resuming from the single-agent checkpoint. Separate models are trained for team sizes of 2, 4, and 8 agents, denoted as CooHOI-2, CooHOI-4, and CooHOI*-8, respectively. CooHOI*-2 converges in roughly 2 days. CooHOI*-4 requires about 6 days, and CooHOI*-8 continues from the 4-agent checkpoint and trains for an additional 5 days. All models are trained with an episode length of 600 timesteps.

\boldparagraph{Contact point assignment} To reduce inter-agent collisions when spawning large teams, we enforce a consistent geometric mapping between agents and their designated contact points.
All contact points are sorted counter-clockwise, starting from the bottom-left corner. After agents are initialized, they are indexed in the same counter-clockwise order, also starting from the bottom-left position. A one-to-one assignment is then performed between agent indices and contact points following this order.

\section{More Experimental Results}

\boldparagraph{Unified policy across all team sizes} To complement the results presented in the main paper, Table~\ref{tab:ours-teamsweep} reports the full performance of our unified policy across all team sizes from 2 to 8 agents under both normal (1$\times$) and heavy (5$\times$) table weights. Beyond the configurations shown in the main paper (2A, 4A, 8A), we additionally include intermediate team sizes (3A, 5A, 6A, 7A), demonstrating that the same decentralized policy generalizes smoothly across all team sizes without retraining. Under the normal-weight setting, our model consistently achieves near-perfect success rates across all team sizes with consistent cooperation. 

Under the heavy-weight setting (5$\times$ table mass), the increased load amplifies the need for coordinated force generation. As team size grows, our unified policy facilitates effective cooperation that leverages the additional mechanical advantage provided by larger groups, resulting in steadily improving success rates with more agents.

\begin{table}[ht!]
\centering
\footnotesize
\setlength{\tabcolsep}{6pt}
\caption{
Performance of our unified policy across team sizes under normal (1$\times$) and heavy (5$\times$) table weights.
}
\label{tab:ours-teamsweep}

\begin{tabular}{ccccc}
\toprule
\multicolumn{5}{c}{\textbf{Normal weight (1$\times$)}} \\
\midrule
\begin{tabular}{c}
\textbf{Team}\\[-2pt]\textbf{size}
\end{tabular} &
\textbf{SR (\%) $\uparrow$} &
\textbf{$d$ (m) $\downarrow$} &
\textbf{$t_\text{coop}$ (\%) $\uparrow$} &
\textbf{$|J|$ (m/s$^3$) $\downarrow$} \\
\midrule
2 & 99.1 & 0.06 & 95.2 & 51.0 \\
3 & 99.4 & 0.06 & 98.3 & 50.5 \\
4 & 99.2 & 0.08 & 96.1 & 44.7 \\
5 & 99.5 & 0.06 & 97.3 & 40.6 \\
6 & 99.3 & 0.07 & 95.9 & 38.0 \\
7 & 98.6 & 0.11 & 93.7 & 35.7 \\
8 & 97.5 & 0.18 & 90.1 & 34.2 \\
\midrule
\multicolumn{5}{c}{\textbf{Heavy weight (5$\times$)}} \\
\midrule
4 & 3.5  & 4.77 & 90.9 & 23.4 \\
5 & 18.2 & 2.48 & 79.0 & 28.3 \\
6 & 50.1 & 1.04 & 79.0 & 32.0 \\
7 & 71.6 & 0.59 & 81.2 & 31.8 \\
8 & 81.1 & 0.49 & 81.5 & 31.7 \\
\bottomrule
\end{tabular}
\end{table}

\begin{table}[ht!]
\centering
\footnotesize
\setlength{\tabcolsep}{6pt}
\caption{
Performance of our unified policy across team sizes for small and large tables. All results are averaged over 10,000 simulation episodes.
}
\label{tab:ours-small-large}

\begin{tabular}{ccccc}
\toprule
\multicolumn{5}{c}{\textbf{Small tables}} \\
\midrule
\begin{tabular}{c}
\textbf{Team}\\[-2pt]\textbf{size}
\end{tabular} &
\textbf{SR (\%) $\uparrow$} &
\textbf{$d$ (m) $\downarrow$} &
\textbf{$t_\text{coop}$ (\%) $\uparrow$} &
\textbf{$|J|$ (m/s$^3$) $\downarrow$} \\
\midrule
2  & 93.1 & 0.37 & 94.8 & 63.2 \\
3  & 97.5 & 0.14 & 97.0 & 64.7 \\
4  & 98.4 & 0.12 & 97.1 & 55.5 \\
8  & 96.4 & 0.23 & 85.4 & 45.0 \\
\midrule
\multicolumn{5}{c}{\textbf{Large tables}} \\
\midrule
2  & 71.0 & 0.85 & 91.7 & 53.3 \\
3  & 82.7 & 0.46 & 94.3 & 52.1 \\
4  & 85.3 & 0.51 & 94.6 & 48.8 \\
8  & 84.2 & 0.93 & 86.1 & 45.4 \\
12 & 80.6 & 1.14 & 58.2 & 45.2 \\
16 & 74.5 & 1.41 & 15.1 & 46.6 \\
\bottomrule
\end{tabular}
\end{table}

\boldparagraph{Zero-shot generalization}
We further evaluate our unified policy under unseen table geometries and team sizes, testing whether the coordinated formation and carrying skills acquired during training transfer to new scenarios. We consider both smaller tables (round with $1.40 \text{ m}$ diameter, square $1.30 \text{ m} \times 1.30 \text{ m}$, rectangular $1.60 \text{ m} \times 0.90 \text{ m}$) and larger tables (round with $2.40 \text{ m}$ diameter, square $2.20 \text{ m} \times 2.20 \text{ m}$, rectangular $3.0 \text{ m} \times 1.40 \text{ m}$), all with the same mass density as in the main experiment.

As shown in Table~\ref{tab:ours-small-large}, our policy maintains coherent cooperation across all configurations despite this distribution shift. For smaller tables, agents occasionally display slightly stronger lift initiation, resulting in modestly higher jerk, but the transport phase remains stable and success rates stay consistently high. For larger tables, agents still maintain synchronized cooperative behaviors. However, the increased mass and longer moment arms make lifting and stabilizing harder. Consequently, agents can sometimes lose balance, fall, and trigger early termination. The large-table setting is particularly more challenging for two-agent teams, which have less mechanical leverage to stabilize and lift the heavier tables, resulting in slower transport.

\begin{figure}[H]
    \centering
    \includegraphics[width=1.0\linewidth]{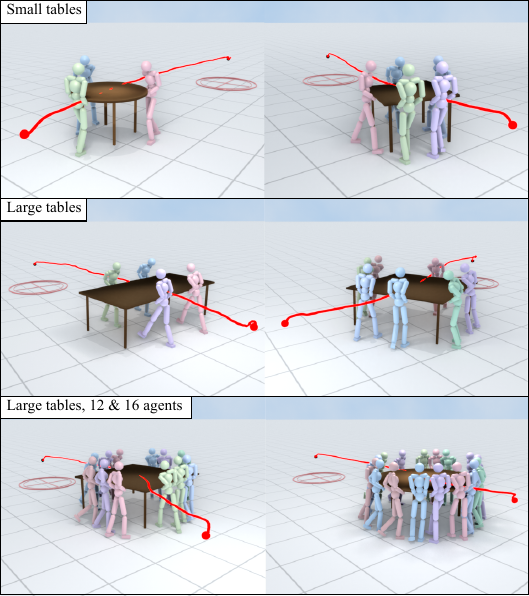}
    \caption{Qualitative visualization of the zero-shot generalization under unseen table geometries and team sizes. Red line indicates the table’s movement trajectory, and the black dot marks its final position at the end of each episode.
    }
    \label{fig:zeroshot}
\end{figure}

We also evaluate zero-shot generalization to 12-agent and 16-agent teams carrying the large tables, pushing the policy far beyond the team sizes encountered during training. The unified policy continues to produce synchronized and coherent motion, achieving relatively high success rates and low jerk, in contrast to the baseline which becomes highly unstable. However, when teams become very large, the tabletop perimeter becomes crowded, and agents have not fully learned to position themselves within tight support gaps, resulting in lower cooperative-time ratios. Nonetheless, our policy exhibits overall robust generalization to object sizes and larger team sizes unseen during training. We provide several qualitative results in Figure~\ref{fig:zeroshot}.

\vspace{5cm}

\section{Multiple Affordance Behaviors}

\begin{figure}[H]
\centering
\includegraphics[width=0.48\linewidth]{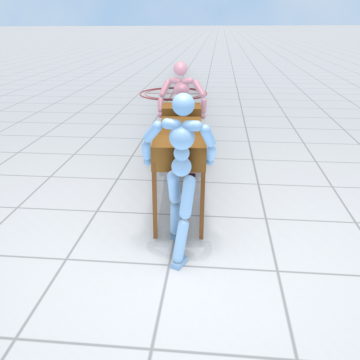}
\hfill
\includegraphics[width=0.48\linewidth]{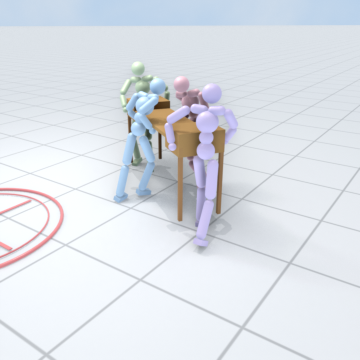}
\caption{Examples of multiple affordance behaviors learned by adapting the task reward. The agents are able to adapt to side-holding or edge-lifting while being able to walk toward diverse directions. The policy is trained using the same single-human reference motions as our main experiments.}
\label{fig:affordance}
\end{figure}

While our main experiments focus on a single affordance behavior (edge-lifting), our framework can support multiple affordance behaviors by adapting the task reward. Fig.~\ref{fig:affordance} demonstrates this capability, where agents adapt to either side-holding or edge-lifting depending on their proximity to regions where the corresponding affordances are feasible.

\end{document}